\documentclass[twoside]{article}

\usepackage{color,soul} 
\usepackage[pdftex]{graphicx}
\usepackage[accepted]{aistats2020}

\usepackage{hyperref}

\usepackage{color,soul}

\usepackage{amssymb}
\usepackage{xcolor}
\usepackage{algorithm}
\usepackage{algorithmic}
\usepackage{pgfplots}
\usepackage{multirow}
\usepackage{array}
\usepackage{graphicx}
\usepackage{subcaption}
\usepackage[square]{natbib}
\usepackage{diagbox, eqparbox, hhline}
\begin{document}

%
\runningtitle{Uncertainty Quantification for Deep Context-Aware Mobile Activity Recognition}

%
\runningauthor{Z. Huo, A. Pakbin, X. Chen, N. Hurley, Y. Yuan, X. Qian, Z. Wang, S. Huang, B. Mortazavi}

\twocolumn[

\aistatstitle{Uncertainty Quantification for Deep Context-Aware Mobile Activity Recognition and Unknown Context Discovery}

\aistatsauthor{Zepeng Huo$^{1}$ \And Arash Pakbin$^{1}$ \And  Xiaohan Chen$^{1}$ \And Nathan Hurley$^{1}$ \And Ye Yuan$^{1}$}

\aistatsauthor{Xiaoning Qian$^{1}$ \And Zhangyang Wang$^{1}$ \And Shuai Huang$^{2}$ \And Bobak J. Mortazavi$^{1}$ }

\aistatsaddress{ Texas A\&M University$^{1}$ \And  University of Washington$^{2}$} ]

\begin{abstract}
Activity recognition in wearable computing faces two key challenges: i) activity characteristics may be context-dependent and change under different contexts or situations; ii) unknown contexts and activities may occur from time to time, requiring flexibility and adaptability of the algorithm. We develop a context-aware mixture of deep models termed the $\alpha$-$\beta$ network coupled with uncertainty quantification (UQ) based upon maximum entropy to enhance human activity recognition performance. We improve accuracy and F score by 10\% by identifying high-level contexts in a data-driven way to guide model development. In order to ensure training stability, we have used a clustering-based pre-training in both public and in-house datasets, demonstrating improved accuracy through unknown context discovery.
\end{abstract}

\section{Introduction}\label{intro}

\begin{figure*}[!ht]
\centering
\includegraphics[width=2\columnwidth]{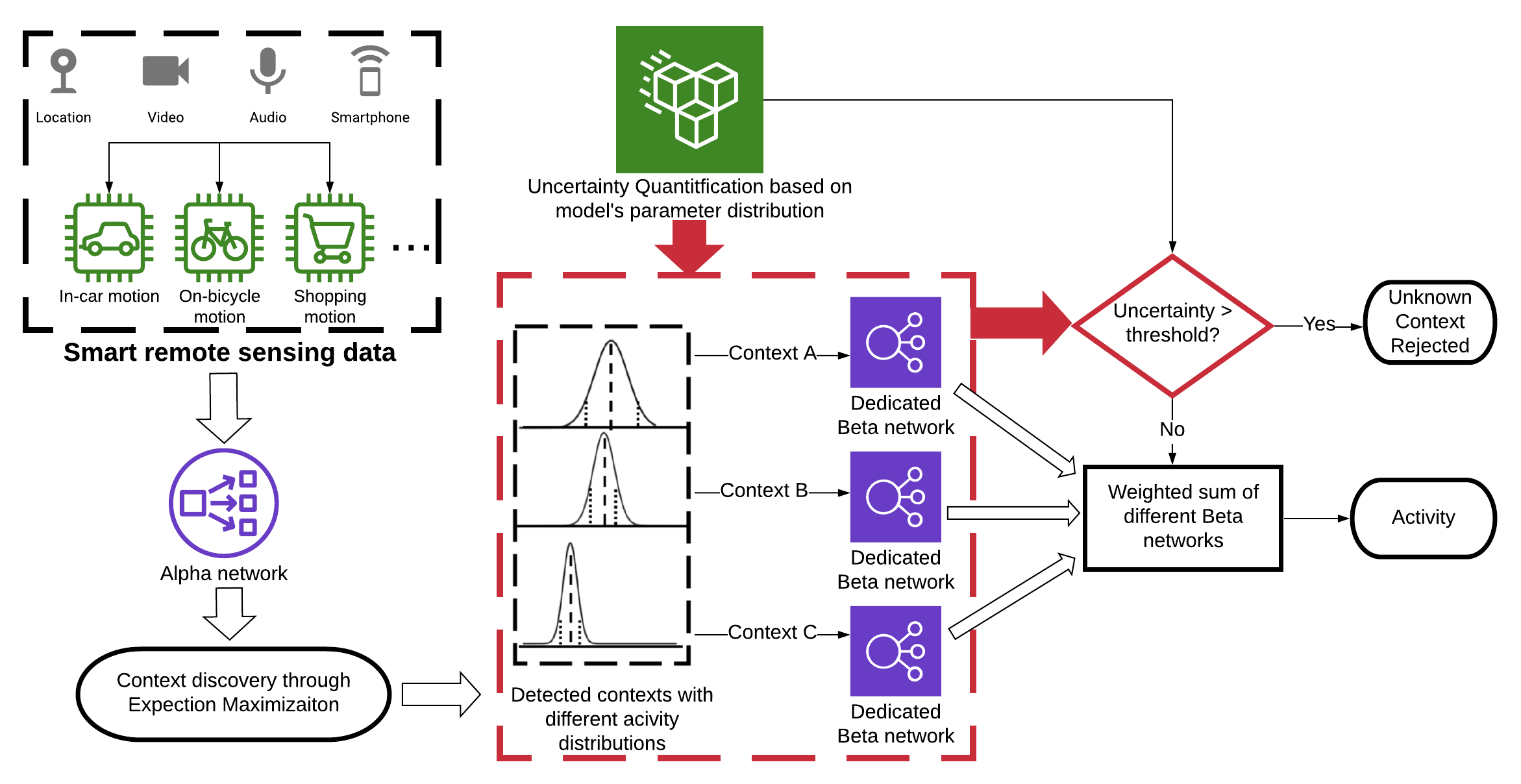}
\caption{A conceptual overview of UQ integrated with the $\alpha$-$\beta$ network}
\label{fig:pipeline_flowchart}
\end{figure*}

In wearable computing, context-awareness helps recognize activities based on sensor measurements under different situations. Context can be defined as ``any information that can be used to characterize the situation'' \citep{sezer2018context} or to improve recognition \citep{context-definition}. Context-aware systems have previously been used in many applications, including activity recognition \citep{context-aware-activity-recognition-cosar}, online, personalized and adaptive activity classification \citep{xu2016personalized}, and healthcare applications \citep{andreu2015wearable,spiegel2014validation}.
The definition of context heavily relies on domain knowledge, such as a user's tasks (e.g., spontaneous activity, engaged tasks) or a user's social environment (e.g., co-location of others, group dynamics), etc.  However, in practice, pre-defined contexts may not always be available, or definitions of contexts may change in different environments. Additionally, new unknown contexts may emerge over time. For these reasons, there is a general data insufficiency and lack of contextual information to develop accurate context-aware activity recognition systems that could adapt to these unknown contexts. 
While existing works are not adequate to address the challenges such as a lack of context information in data while training the model, or the emergence of unknown contexts when the model is applied \citep{context-driven, steven2018feature}, we propose a data-driven approach with context-awareness capability to achieve better activity recognition performance. Specifically, we develop an integrative $\alpha$-$\beta$ framework to simultaneously learn unknown contexts and the distribution of each user's specific activity likelihood within each context. 
In this framework, the $\alpha$ network is the context detector to learn a distribution over contexts as a mixture of weights, and the $\beta$ network models activity recognition for context-specific sensor measurements. For example, given the sensor reading data from a user, the $\alpha$ network detects the context by generating a distribution over different contexts; then, each context has a dedicated $\beta$ network that outputs a distribution over different activities. 

We further extend our model with the ability to explore new unknown contexts by equipping the $\alpha$-$\beta$ network with uncertainty quantification (UQ) based on the maximum entropy learning (MEL) principal. MEL identifies the distribution of the parameters of a statistical model that bears the maximum uncertainty, rather than one single best model, as a principle to achieve robustness in prediction and modeling. The prediction model could refuse to predict on given data if the uncertainty for making a prediction on this data is higher than a threshold. This method adapts to data and effectively discovers unknown contexts with the UQ. This work allows for models trained in laboratory settings to extend to natural environments for monitoring behaviors and performances of users, as in Figure \ref{fig:pipeline_flowchart}.

\textcolor{black}{Our contributions are as follows. In this paper, we propose a context-aware model for activity recognition. The context and activity are simultaneously modeled by dedicated networks. For unknown contexts, an overarching UQ method is applied to all the model parameters. This provides robustness in testing new context that our model can uniquely offer, beyond a traditional activity recognition technique}. Finally, we demonstrate these findings in both a publicly-available benchmark dataset and an in-house dataset we collected to identify confounded versions of human motion, and make this data available for public use.

\section{Related Work}\label{related_work}
\textbf{Context-Awareness: Mixture of Experts Model}
In many applications of machine learning, heterogeneous data can be divided into smaller homogeneous groups that can be modeled more accurately. Context-awareness, as an example, plays an important role in improving the performance of activity recognition systems \citep{ravi2005activity,wang2019deep}. In Mixture of Experts (MoE) models \citep{jordan1994hierarchical} such group-level clustering and modeling comes as a single training step, rather than splitting data a prior then building models. Elements are clustered based upon their relationship and the next level modeling accuracy. MoEs have successfully served different applications from classification and regression tasks \citep{yuksel2012twenty}\citep{miech2017learnable} to phenotyping in medical datasets \citep{courbariaux2018mixture}. 
In \citep{jordan1994hierarchical}\citep{yuksel2012twenty}\citep{jordan1995convergence}, authors provided formulations of probabilities of observed given different experts.
The MoE structure consists of two components: A gate and several experts. The gate is often modeled by Gaussian Mixture Models (GMM) \citep{yuan2009variational}\citep{sharma2019flexible} and neural networks \citep{lima2007hybridizing}, while expert modeling is more dependant on the application including SVM \citep{lima2007hybridizing} \citep{cao2003support}. This work used neural networks for both.

Different methods have been used in the literature in order to determine the number of experts for these models \citep{yuksel2012twenty}: growing models where the experts with the worst performance decompose into a MoE themselves \citep{shazeer2017outrageously} \citep{aljundi2017expert}, pruning models where they start with a large number of experts and reduce the number by combining/removing experts \citep{jacobs1997bayesian}, and exhaustive search in cases where tree topologies are not overly complex \citep{bishop2002bayesian}.



\color{black}

\textbf{Uncertainty Quantification: Previous Studies}
UQ has been critical for robust learning under different contexts (known or unknown) in mobile activity recognition  \citep{ardywibowo2019adaptive}, healthcare \citep{samareh2019uq,meghdadi2017brain}, signal processing \citep{reynders2016uncertainty}, and manufacturing \citep{nannapaneni2014uncertainty}. Existing models with Gaussian process \citep{ardywibowo2019adaptive} or Bayesian approximation methods \citep{gal2016dropout} either rely on the assumption that variables in a system can be characterized by explicit probabilistic relationships (e.g., Bayesian models) or rely on generating one best model in the learning algorithm. However, measuring the predictive uncertainty for deep neural networks remains a challenging problem, closely related to the problem of detecting test samples that are drawn sufficiently far away from the distribution of training samples. 
For example, Malinin \& Gales proposed a framework called Prior Networks (PNs) for modeling predictive uncertainty that explicitly modeled distributional uncertainty \citep{malinin2018predictive}, Hendrycks \& Gimpel proposed a framework that utilized probabilities from softmax distributions and detected out-of-distribution examples, by introducing confidence scores based on density estimators \citep{hendrycks2016baseline}, later improved by processing the input and output of DNNs in \citet{liang2017principled}. Lee et al. proposed a method for detecting abnormal test samples by including both out-of-distribution and adversarial samples, to obtain the class conditional Gaussian distributions by introducing confidence scores based on the Mahalanobis distance \citep{lee2018simple}. 
Our method is different from these works as we aim to learn a distribution of deep models rather than one single best one.

\section{Methods}\label{Methods}
Contextual information can help model similar activities together, resulting in an improvement of activity recognition performance by reducing the search space of activities to recognize given a set of features.
Our proposed $\alpha$-$\beta$ network integrates activity recognition with unsupervised context detection, as detailed in Section \ref{background}. As context can vary from person to person, and change over time for the same person, 
 in Section \ref{discovery}, we present maximum entropy learning (MEL) based UQ in the $\alpha$-$\beta$ network to discover unknown contexts when needed. 

\subsection{Context-Awareness Processing}\label{background} 
\normalsize
In this work we develop a mixture of CNNs, the $\alpha$-$\beta$ network, where each mixture component is dedicated to one specific context. There are two types of networks: $\alpha$ and $\beta$. Given the sensor data, the $\alpha$ network detects context by generating a probability distribution over all known contexts. Each context has a dedicated $\beta$ network that outputs a probability distribution over different activities. Our activity recognition problem features a latent context variable and can be formulated as:

\noindent
\footnotesize
\begin{equation}\label{eq1}
\begin{aligned}
&\log {p(ACTIVITY|\mathbf{X}, \mathbf{\theta})} = \sum_{i=1}^{N}{\log {p(activity_i|\mathbf{x}_i, \mathbf{\theta})}} \\
&= \sum_{i=1}^{N}{\log {\sum_{c=1}^{N_c}{p(activity_i| c_i=c, \mathbf{x}_i, \theta)p(c_i=c| \mathbf{x}_i, \mathbf{\theta})}}},
\end{aligned}
\end{equation} 
\normalsize

where $\theta$ denotes the mixture component parameters, $N$ denotes the number of data samples, and  $N_c$  denotes the number of expected clusters (contexts) to which each data point may belong. Our objective is to maximize Eq. \eqref{eq1} with respect to $\theta$. As shown by \citep{dempster1977maximum}, the log-likelihood has a lower bound, first formulated in \citep{solis2019human}:  
 
\footnotesize
\begin{equation}
\begin{aligned}
&{\log {p(ACTIVITY|\mathbf{X}, \mathbf{\theta})}} \geq \label{eq2}\\
&\sum_{n=1}^{N}{\sum_{c=1}^{N_c}q(c_i=c)\log{\frac{p(activity_i|c_i=c, \mathbf{x}_i,\mathbf{\theta}).p(c_i=c)}{q(c_i=c)}}},\nonumber 
\end{aligned}
\end{equation}
\normalsize

\noindent where $q(\cdot)$ is the distribution over different contexts and $p(\cdot|context, x, \theta)$ is the distribution over activities given the context and the input. $q(\cdot)$ is modeled using the context-detecting $\alpha$ network, and each $p(\cdot|context, x, \theta)$ is modeled using a $\beta$ network (each context has its own $\beta$ network). While \citep{solis2019human} used a supervised technique to define contexts as specific locations, this work provides for an unsupervised exploration of context. Additionally, in our implementation, we use a network for context recognition ($\alpha$ network) and a dedicated network for each context ($\beta$ network) whereas \citep{solis2019human} used only two networks regardless of the number of contexts. Following the EM algorithm, the lower bound in Eq. \eqref{eq2} can be maximized. Specifically, the loss, the negative of the lower bound, is minimized. In the E-step, $q(\cdot)$ is optimized which translates to optimizing $\alpha$ network while freezing $\beta$ networks. In the M-step, $\theta$ (model parameters) need to be optimized which translates to optimizing $\beta$ networks while freezing $\alpha$ network. The EM training alternates iterations of training either $\alpha$ network or $\beta$ networks while keeping the other(s) fixed. It should be noted that no labeled contextual data is used in the training process for this $\alpha$-$\beta$ network.

\subsection{Unknown Context Discovery}\label{discovery}

The $\alpha$ network enables context detection; however, in practice, contexts may change over time, or may not always be pre-defined. It is possible to improve context-aware systems by detecting the uncertainty of possible unknown contexts as a result of potential distribution mismatch between known and unknown contexts. To identify unknown contexts, we combine the feature extraction power of deep learning with the learning power of MEL to define a probabilistic mechanism for unknown context discovery. 


While many of the current works focus on revising general deep models with a probabilistic evaluation of their model or prediction, here
we have a different aim: We modify the $\alpha$-$\beta$ network to be adaptive to changing contexts hidden in data. We relax our expectation of identifying one single optimal model of the $\alpha$-$\beta$ network; rather, we consider solving for a full distribution over multiple models. The intuition is that many different models might generate relatively similar performance, so it would be better to estimate a distribution over parameters $p(\mathbf{w})$, from the output layer of the $\alpha$ networks that detects context. This aligns with the basic principal of MEL \citep{sensoy2018evidential,bauer2019deep}. Therefore, we equip the $\alpha$-$\beta$ network with UQ capacity based on the MEL principal by identifying the distribution of the parameters of a statistical model that bears the maximum uncertainty. 

\subsubsection{Uncertainty Quantification via Minimizing Relative Entropy} 

To learn the distribution of the $\alpha$-$\beta$ network parameters that encode maximum uncertainty, we employ the MEL formulation. There are two steps. First, we create constraints that encode information from the data. For example, for each sample we derive \textcolor{black}{a loss function} such that the expected prediction on this sample over all the possible model parameters matches the observed outcome on this sample, \textcolor{black}{as in traditional ML}. Second, on the top of this constraint structure, the learning objective of MEL is to learn the distribution of the model parameters with the maximal entropy \textcolor{black}{in terms of the parameter posterior distribution}. Thus, unlike traditional machine learning methods that estimate a single optimal setting of the parameter,  MEL considers a more general problem of these methods by solving for a full distribution over multiple $p(\mathbf{w})$ values.

\subsubsection{Analytical Details of MEL} 
To further illustrate this distribution approach, note that our context detection problem is also a classification problem where the response variable is denoted by $y$ taking values for different contexts. Let $\mathbf{x}_n =[  \mathbf{x}_1,\dots,  \mathbf{x}_n]$
be an input feature vector as an aggregate of all the measures from sensors for each window, and let $\mathcal{D}\left(   \mathbf{x}_{n}|\mathbf{w} \right)$ be the discriminant function parameterized by $\mathbf{w}$ implemented in the $\alpha$ network.
Traditional learning machines such as the max-margin methods estimate the optimal $\mathbf{\hat {w}}$ that minimizes the classification error in predicting the labels of training examples as:
\noindent
\begin{equation}\label{eq:predict}
\hat{y} =sign\mathcal{D}\left(   \mathbf{x}_{n}|\mathbf{w} \right).
\end{equation}
Based on this line of thought, we can classify margin as $y_n\mathcal{D}(\mathbf{x}_n,\mathbf{w})$, and learn the optimal parameter setting $\mathbf{w}$ by the empirical loss and the regularization penalty as:
\noindent
\begin{equation}\label{eq:original theory}
\begin{aligned}
&\min_{(\mathbf{w},\mathbf{\gamma}_n)} R(\mathbf{w}) + \sum_{n} L(\mathbf{\gamma}_n)\\
&\quad \text{s.t.} \quad y_n\mathcal{D}(  \mathbf{x}_n\mid\mathbf{w}) - \mathbf{\gamma}_n
 \geq \mathbf{0}, \quad \forall n. 
 \end{aligned}
\end{equation}
where $L(\mathbf{\gamma}_n)$ is the loss function, a non-increasing and convex function of the margin, and $R(\mathbf{w})$ is the regularization penalty.
Given $p(\mathbf{w})$, we can recast \eqref{eq:original theory} as an integration where the classification constraints will also be applied in an expected sense. Instead of considering an expectation of the regularization penalty functions, we can apply a canonical penalty function for distributions, the negative entropy; minimizing the negative entropy is equivalent to maximizing the entropy. Hence, we use the Shannon entropy defined as $H(p(\mathbf{w})) = - \int p(\mathbf{w})\log p(\mathbf{w})d\mathbf{w}$. This gives us the following objective function to learn the distribution $p(\mathbf{w})$ over the parameters $\mathbf{w}$:
\begin{equation}\label{eq:MEL objective}
 \begin{aligned}
&\min_{p(\mathbf{w})} H(p(\mathbf{w})) \\
&\quad \text{s.t.} \quad \int p(\mathbf{w})[y_n\mathcal{D}(  \mathbf{x}_n,\mathbf{w}) - \gamma_n]d\mathbf{w} \geq \mathbf{0}, \quad \forall n.
 \end{aligned}
\end{equation}
As a result, MEL no longer finds a fixed set of the parameters, but a distribution over them. Learning such a distribution of model parameters does not rely on assumptions on the model's mathematical form.  It also does not rely on knowing a particular distribution as is needed in Bayesian learning frameworks. Therefore, MEL is more flexible than typical Bayesian learning methods \citep{sun2018multi,zhu2018semi} to characterize uncertainties associated with complex models such as the $\alpha$-$\beta$ network here.
To solve the MEL formulation \eqref{eq:MEL objective}, we could derive a Lagrangian $J(p)$, and take the derivatives with respect to $\mathbf{w}$ and set them to $0$. To do that we first need to calculate the unconditional maximum of the problem \eqref{eq:MEL objective} plus the constraints added with some multiplying factors (the Lagrange multipliers), which give the probabilities in a functional form with the Lagrange multipliers as parameters. 
Our UQ approach compares the uncertainty with a threshold to see whether a given sample should be detected as belonging to a new context or not. 
We defined a classification with rejection option as $\hat{y}_i^{Rej}$, where if a sample is rejected $\hat{y}_i^{Rej}=0$, and if it is accepted $\hat{y}_i=\hat{y}$, where $\hat{y}$ corresponds to the classification of the $i$th sample. Note that, a sample is rejected when $p(\mathbf{w}|x_i) < \epsilon$, and $\epsilon$ is chosen through cross-validation.

The distribution of parameters forms a quantitative evaluation of model uncertainty, which could be further used in subsequent decision making by using probability laws to track the uncertainty propagation process. In this paper, we create a rejection option, a flexibility enabled by the UQ capacity. The rejection option allows for the prediction model to refuse to generate a prediction if the uncertainty is higher than a given threshold. This is typically solved by estimating the class conditional probabilities and rejecting the samples that have  lower posterior probability of class. 

\section{Experiments}\label{experiements}

In this section, we discuss experiments with our deep context-aware mixture of experts for activity detection coupled with maximum entropy based uncertainty quantification. We first introduce the datasets, and then the detailed implementation of our models. Then, we show various competitive baselines to demonstrate that our pipeline performs the best as evaluated by several different metrics. 
\subsection{Datasets}

\textbf{UCI data.} We used the UCI OPPORTUNITY dataset \citep{roggen2010collecting} for context-aware human activity recognition. The dataset contained 18 different activities performed in five different contexts and sensed by 72 different sensors. Each of the 18 activities had one of the five contexts, but not all contexts contained every activity. Therefore, the UCI OPPORTUNITY dataset provided a realistic capture of a situation where not all of the human activities occurred with an equal likelihood in all contexts. The UCI OPPORTUNITY dataset has seven levels of hierarchical labels. Higher level labels described details such as subject posture, while lower level labels described the hand movements or interactions with other subjects. In this study, we chose a higher level label (e.g., cleaning time) as the context and a lower level (e.g., opening a door) label as the activity. 
\textcolor{black}{We used all the body-worn sensors which included seven inertial measurement units (IMUs) and twelve 3D acceleration sensors. Five IMUs were on the upper body while two were on user's shoes. Accelerometers were on the upper body, hip, and leg, which translated to 133 columns in the raw dataset.}

\textbf{In-house data.} To generate a more realistic experimental setting, we have collected our data to detect different types of human motion that may be confounded by environments factors. The dataset is made publicly available, serving as extra part of our contribution in this paper. The motivations are three fold: 1) instead of collecting the data in a strict laboratory setting, we have loosened all the rules for subjects, including their choice of rest time and the pace of activity. 2) we have devised a set of much more realistic contexts to be detected. Those contexts can cover almost all the real world setting a subject might face. 3) we have collected a much nosier dataset compared to previous ones, with respect to the randomness we imposed on the data collection, such as a rare activity as walking with one shoe off in both outdoor (lawn) and indoor (hardwood covered by carpet). 
As a result we have 3 contexts \textcolor{black}{ with corresponding movements} (1) outdoors: Crawling, Jogging, Riding Bike, Sprinting, Walking, Walking with One Shoe (simulated limping), Walking with Weight in Arms, Walking with Weight on Back. 
(2) Movements that happen indoors: Escalator Up, Escalator Down, Elevator Up, Elevator Down, Lying Down, Sitting, Stairs Down, Stairs Up, Standing, Walking, Walk with One Shoe (simulated limping), Cooking, Dancing, Eating, Reading, Sleeping, Talking on phone, Talking to Another Person, Using PC, and 
(3) movements that happen outdoors but in vehicles: Driving Car, Riding Car, Riding Bus, Reading, Device Usage.
The movements, however, are not unique to each context, and the position of the phone may change through usage. We collected data on 20 people, each doing 32 activities while having 3 phones, one in hand, one in the pocket and one in the backpack\textcolor{black}{, as those are the common places for phone positions}. The applications used on the phones for data collection was \textit{Readisens} \citep{ReadiSens}. \textcolor{black}{The sensors used are: acceleration  (in three axes), altitude, compass, gyroscope (in three axes), GPS information (latitude, longitude), screen time information, and phone speed.} This data contains contextual information which is independent of the locations of the phone. Participants were given minimal instruction for the execution of the activities in order to allow for individual variation. Participants were also wearing clothes of choice and the order of the activities were randomized. \textcolor{black}{This study was reviewed and approved by the Texas A\&M Institutional Review Board (IRB \# 2018-210D). The data has been made available for public use.\footnote{https://github.tamu.edu/guangzhou92/RealActivity}}

\color{black}


\subsection{Implementation}

In UCI dataset, 19 different preprocessed sensors were fed into the network. Time series data were divided into non-overlapping segments of 1 second (30 samples). \textcolor{black}{In each window, the features of sensors are concatenated.}
We used a five-fold cross-validation to evaluate our models with testing accuracy and micro F score as performance metrics. In our experiments, we used 3 convolutional layers followed by 3 fully connected layers for both $\alpha$ and $\beta$ networks. Note that these two networks' architectures were different in terms of the number of neurons in the output layer. Networks were trained using stochastic gradient descent with an initial learning rate of 0.001 and a momentum of 0.9, which provided the best results in cross-validation.

\textbf{Implementation of MEL.} For maximum entropy classifier, we have the input from the parameter distribution derived from the $\alpha$ network. This network outputs $\hat{y}_i^{Rej}$ as a probability of rejection, which is compared against a threshold that is derived from cross-validation to further guide the $\beta$ network for fine-grained activity detection under specific context.
We defined a classification with rejection, where if a sample was rejected (if the uncertainty was higher than a specific threshold), the prediction model refused to generate a prediction by setting the predicted context to zero.

In the UCI OPPORTUNITY dataset we had five contexts: relaxing, coffee time, early morning, clean-up, and meal time. To test our unknown context discovery, we adopted a rotating strategy. In this strategy, we removed one context and its corresponding data from the training dataset at each rotation, and trained an $\alpha-\beta$ network only on the remaining known contexts.
\textcolor{black}{In this case, one context was assumed to be unknown and was treated as a hold-out to be used for unknown context discovery assessment.} The final evaluation was conducted through comparison of activities that were sampled from both known and unknown contexts, to demonstrate the model's ability to distinguish the unknown context.

\textbf{Pre-training.}
\label{experiments:pretraining}
Initialization is an essential step for both the optimization and for the training of the neural networks. Without proper initialization, the model collapses into selecting only one specific $\beta$ network while eliminating the contribution of others, \textcolor{black}{as is observed in many other mixture network modeling}. We have added pre-training to solve this problem and have compared it with regularization. We have shown that it is much more effective in terms of accuracy. Pre-training is an important stage in the $\alpha$-$\beta$ network training. The model could approach the base model of a single neural network classifier without proper pre-training because of the large gradients at the beginning of the training stage; large gradients cause the selector to saturate and select only one $\beta$ network. Therefore, the full capabilities of the $\alpha$-$\beta$ network can only be achieved using proper pre-training, which proves useful in finding subgroups of data as well as in yielding a better performance. We used the idea presented in \citep{guerin2017cnn} to cluster the activity data. In detail, a base network was trained with sensor readings as input and activities as output. 
Next, the CNN segment of the network was used to embed the sensor readings into the features which have proven to be descriptive of the input data \citep{sharif2014cnn, donahue2014decaf}. Subsequently, we used K-means to cluster the input into a fixed number of clusters. Finally, we trained our $\alpha$ network to learn the mapping between sensor readings of activity to clusters.


\begin{figure*}[ht!]
\vskip 0.2in
\begin{center}$
\begin{array}{lll}
\includegraphics[width=45mm]{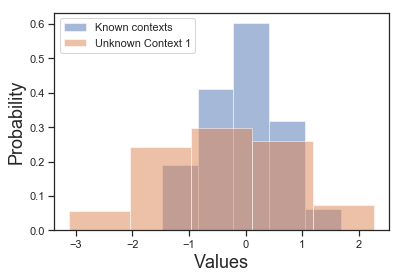}&
\includegraphics[width=45mm]{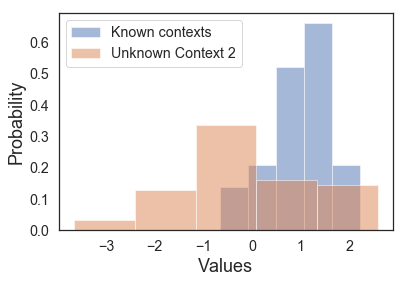}&
\includegraphics[width=45mm]{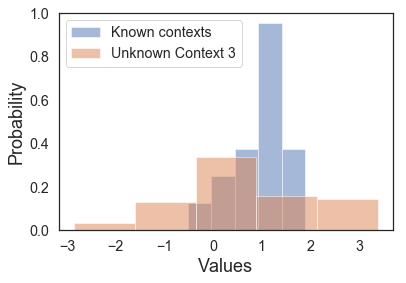}
\end{array}$
\end{center}

\begin{center}$
\begin{array}{rr}
\includegraphics[width=45mm]{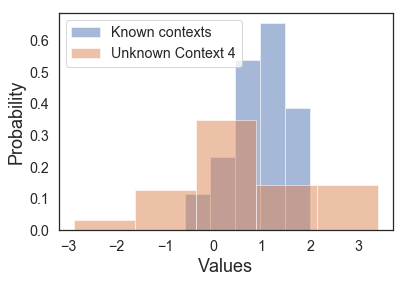}&
\includegraphics[width=45mm]{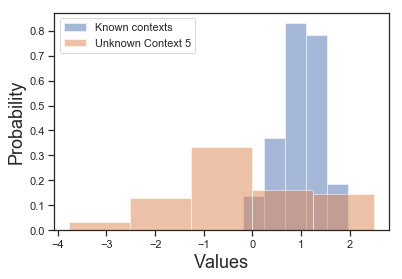}
\end{array}$

\end{center}
\caption{Predicted probability distributions when a context is removed v.s the aggregate of all known contexts within each rotation.}
\label{fig:uncertVScert}
\vskip 0.1in
\end{figure*} 

\subsection{Baseline}
\label{experiments:baseline}

We compare our model against several baselines. The first baseline is a single $\beta$ network, which is a component of the $\alpha$-$\beta$ network. Another baseline, for each specific number of contexts, is a $\beta$ network which has wider hidden layers in order to have the number of parameters equal to the $\alpha$-$\beta$ output (similar size). In other words, for a $\alpha$-$\beta$ network with the number of clusters equal to $k$, given that $\alpha$ and $\beta$ networks have the same size, the baseline is a $\beta$ network with hidden layers $k+1$ times as large to have roughly the same capacity. Finally, 
\textcolor{black}{in order to demonstrate the performance of our UQ method, we design a similar probabilistic baseline, logistic regression, to output a rejection likelihood and we compare it with our method to show the better unknown context discovery performance from our UQ pipeline. Baseline doesn't impose extra regularization, and all parameters were selected through cross-validation.}
\color{black}

\subsection{Result of UCI dataset: UQ vs. Baseline}
\textcolor{black}{For the performance of UQ against baseline, results are shown in Table \ref{table:UQ}.} As can be seen, the proposed UQ method detects unknown contexts better than baseline in all evaluations. 

\noindent
\begin{table}[t]
\caption{UQ results VS. baseline for unknown context discovery where contexts are 1 = Relaxing, 2 = Coffee time, 3 = Early morning, 4 = Clean up, 5 = Sandwich time} 
\label{table:UQ}
\vskip 0.15in
\begin{center}
\begin{small}
\begin{tabular}{llllll}
\hline
\multicolumn{1}{l}{\begin{tabular}[l]{@{}l@{}}Performance \\ measure \end{tabular}} & \multicolumn{5}{c}{\begin{tabular}[c]{@{}c@{}}Context number that \\ was removed each rotation\end{tabular}} \\ \cline{2-6} 
\multicolumn{1}{c}{} & 1 & 2 & 3 & 4 & 5 \\ \hline
\multicolumn{6}{c}{\textbf{UQ}} \\
Sensitivity &0.63  & 0.71 & 0.75 & 0.62 & 0.73  \\ 
Specificity & 0.72 &0.72  &0.76  & 0.79 & 0.82 \\ 
Testing accuracy & 0.67 & 0.71 & 0.75 & 0.69 & 0.77 \\ 
F-score & 0.67 & 0.72 & 0.77 &  0.70& 0.78 \\ \hline
\multicolumn{6}{c}{\textbf{Baseline}} \\
Sensitivity & 0.26 & 0.34 & 0.17 & 0.19 & 0.26  \\ 
Specificity & 0.75 & 0.79  & 0.81  & 0.80 & 0.80 \\ 
Testing accuracy & 0.66 & 0.72 & 0.64 & 0.64 & 0.69 \\ 
F-score & 0.26 & 0.34 & 0.17 & 0.20 & 0.27 \\ \hline
\end{tabular}
\end{small}
\end{center}
\vskip -0.1in
\end{table}
\normalsize
\vspace{-20pt}


 
Figure \ref{fig:uncertVScert} presents a main result of UQ in this experiment from the UCI OPPORTUNITY dataset.  Here, the distribution of predicted probabilities of the removed context (the red histogram) is shown with the distribution for all the other contexts combined (the blue histogram). The UQ algorithm is able to detect the uncertainty for the unknown context, as the unknown context usually leads to smaller probabilities in comparison with the distribution of the known contexts. 

\normalsize

\subsection{Results of UCI dataset: $\alpha$-$\beta$ Network vs. Baseline}
\label{experiments:mix_vs_baseline}

\begin{table}[t!]

\caption{Accuracy (F score) for the $\alpha$-$\beta$ network and baseline which is a $\beta$ network with the same number of parameters for each specific number of clusters. For $1$ cluster both the models are the same.}
\label{tab:AccuracyFscore}
\vskip 0.15in
\begin{center}
\begin{small}
\begin{tabular}{cccc}
\hline
Clusters    & 2 & 3 & 4\\
\hline
$\alpha$-$\beta$ & \textbf{0.89(0.90)} & \textbf{0.89(0.90)} & \textbf{0.91(0.91)} \\
baseline & 0.81(0.81) & 0.84(0.84) & 0.84(0.83) \\
         &                        &                        &\\
\hline
Clusters    & 5 & 6 & 7 \\
\hline
$\alpha$-$\beta$ & \textbf{0.92(0.91)} & \textbf{0.91(0.92)} & \textbf{0.96(0.96)}\\
baseline & 0.83(0.82)                   & 0.86(0.86)                    & 0.85(0.85) \\
         &                        &                        &\\
\hline
Clusters  & 8 & 9 & 10\\
\hline
$\alpha$-$\beta$ & \textbf{0.96(0.96)} & \textbf{0.96(0.96)} & \textbf{0.95(0.95)}\\
baseline & 0.84(0.84) & 0.86(0.86) & 0.86(0.86) \\   

\end{tabular}
\end{small}
\end{center}
\vskip -0.1in
\end{table}

\color{black}


Table \ref{tab:AccuracyFscore} compares the testing accuracy of the  $\alpha$-$\beta$ network and the baseline (a single network that is equivalent to a single $\beta$ network) for predicting labels in the UCI OPPORTUNITY dataset. The testing accuracies are averaged among all their corresponding bootstraps (bootstrapped 5 times).  Table \ref{tab:AccuracyFscore} shows that the mixture of the context-specific neural networks improved on the accuracy of the baseline from 86\% to 96\% when using nine contexts. Thus, our context-aware $\alpha$-$\beta$ network was able to find subgroups in the data in an unsupervised manner with different numbers of contexts
This fact was reflected in the accuracy boost due to the subgroup modeling by different $\beta$ networks. 

\begin{figure}[t!]  
\begin{minipage}[b]{0.48\linewidth}
  \centering
  \centerline{\includegraphics[width=4cm]{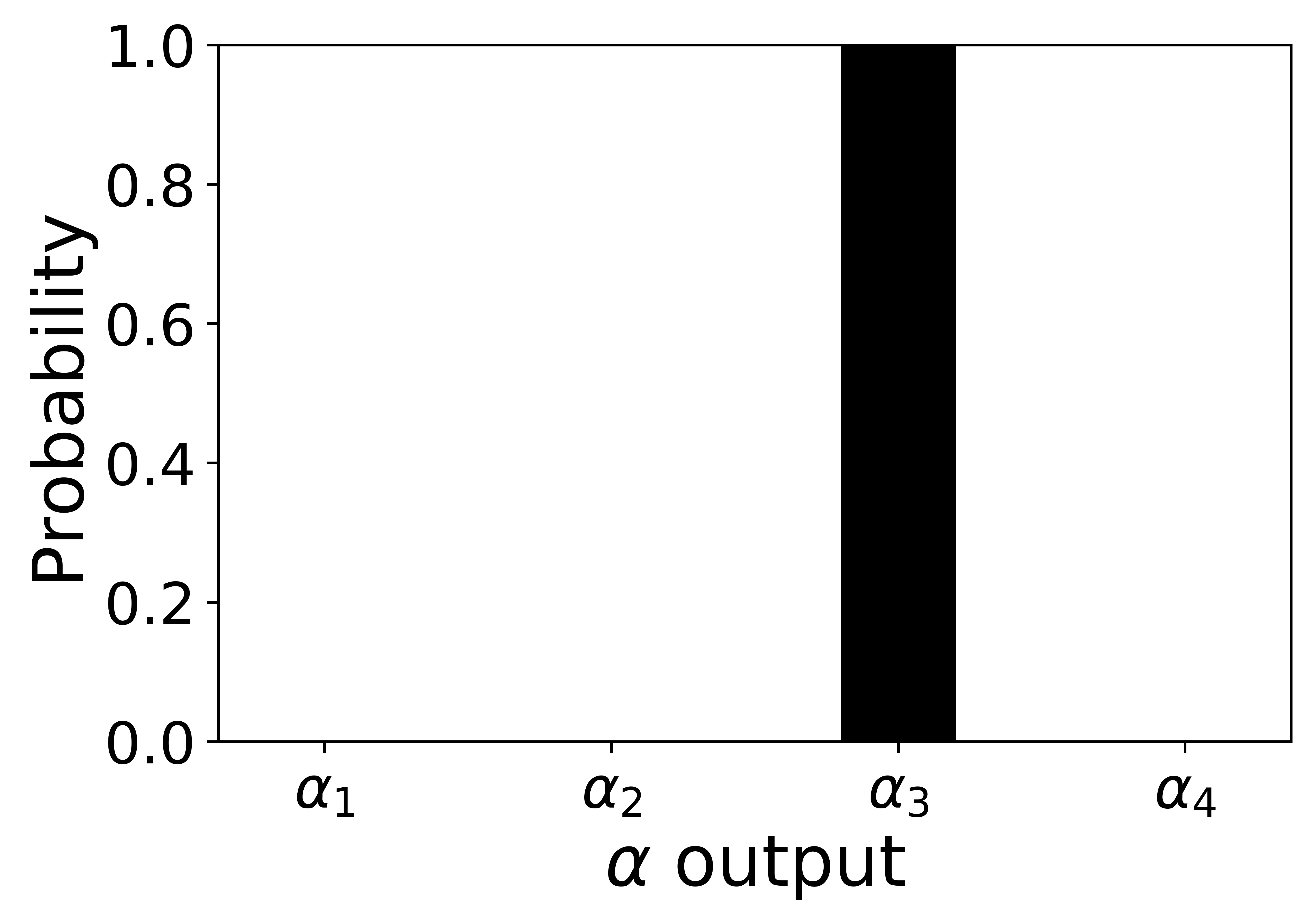}}
  \centerline{(a) Without pre-training}\medskip
\end{minipage}
\hfill
\begin{minipage}[b]{.48\linewidth}
  \centering
  \centerline{\includegraphics[width=4cm]{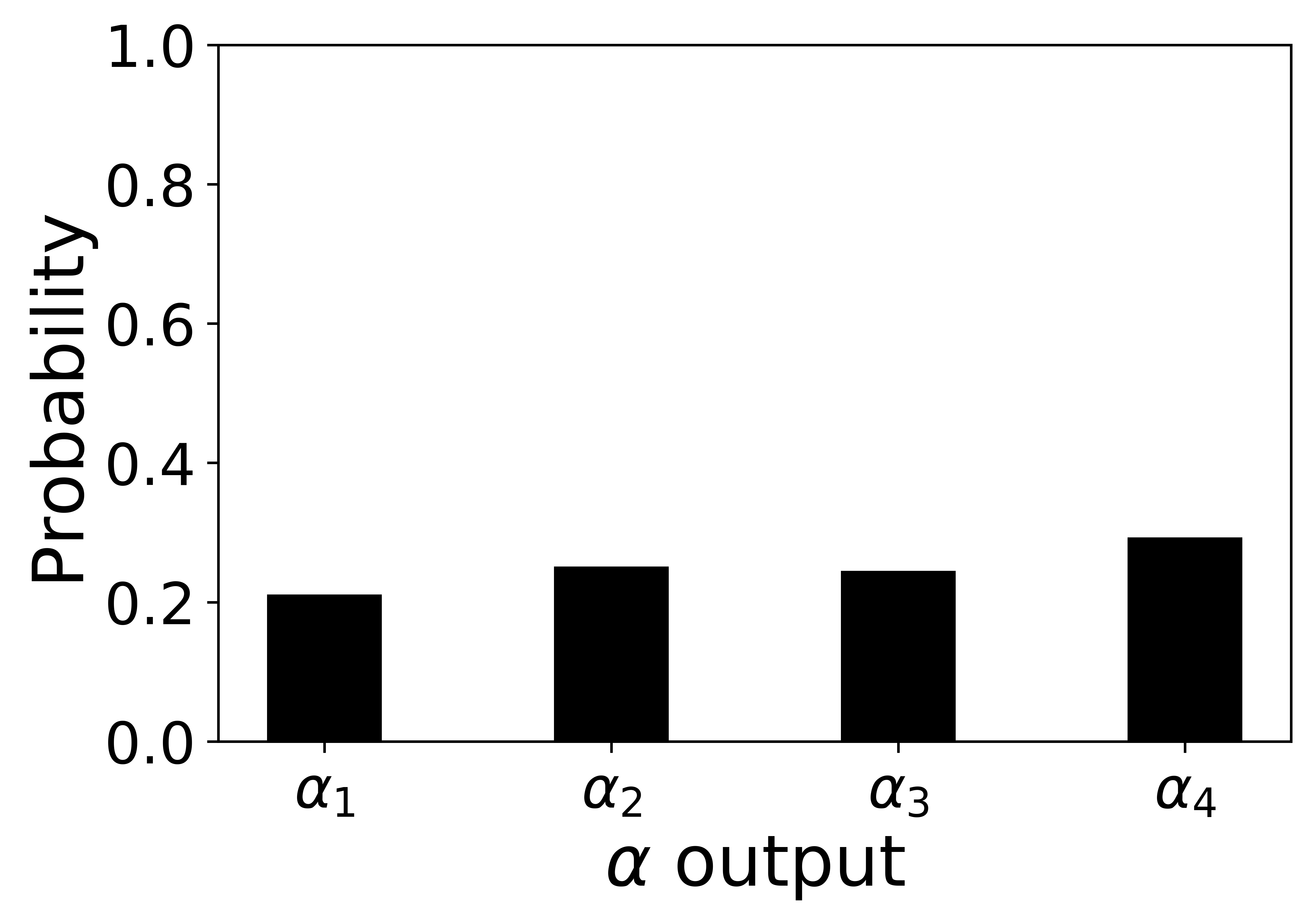}}
  \centerline{(b) With pre-training}\medskip
\end{minipage}
\caption{Average of $\alpha$ network output in testing without (a) and with (b) pre-training. The model collapses into a single network in the former. This shows the effectiveness of pre-training in making sure that the $\alpha$ network finds subgroups in the data and hence can take advantage of context-aware recognition.}
\label{fig:alpha_output}
\end{figure}

\begin{figure}[t!]  
\centering
\includegraphics[width=6.5cm]{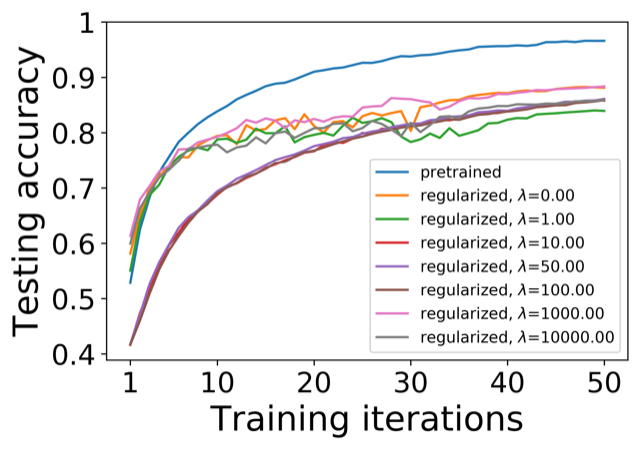}
\caption{Comparison between pre-training and regularization.}
\label{fig:pretraining_vs_regul}
\end{figure}

\textcolor{black}{The $\alpha$-$\beta$ network is equipped with pre-training. The comparison in Figure \ref{fig:alpha_output} shows that pre-training is extremely important, as the model without pre-training results in selection of only one network, leading to model collapse. 
The effect of pre-training can also be seen in the network performance. Figure \ref{fig:pretraining_vs_regul} shows that pre-training is much more effective than regularization since regularization just penalizes single network usage but does not use any knowledge about the data in the process. 
We took the best number of contexts in terms of accuracy and F-score, 9, and tried to train networks with the same number of contexts but without pre-training and by using only regularization with different regularization coefficients.
Without proper initialization, the $\alpha$-$\beta$ network drops from 96\% to 87\% in accuracy and 0.96 to 0.87 in F-score. This is 4\% less than the performance of an identical $\alpha$-$\beta$ network with proper initialization in Table \ref{tab:AccuracyFscore}. This is roughly equal to the performance achieved by a single $\beta$ network (refer to baseline results in Table \ref{tab:AccuracyFscore}).}

\subsection{Results: In-house data}\label{UC_Disc}
\begin{figure}[!t]
\centering
\begin{center}$
\begin{array}{rr}
\includegraphics[width=40mm]{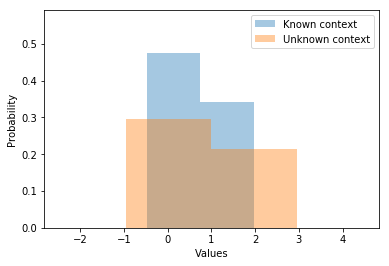}&
\includegraphics[width=40mm]{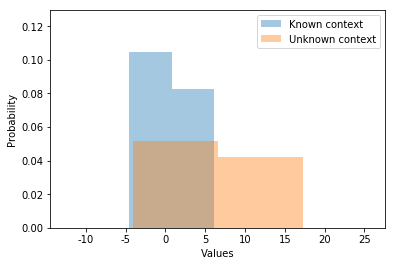}
\end{array}$
\end{center}
\caption{Predicted probability distributions when 1 context is removed vs. 2 contexts are removed 
}
\label{fig:inhouse_UQ}
\end{figure}

Having tested our methodology on the UCI OPPORTINITY dataset, we use our In-house dataset to further assess $\alpha$-$\beta$ Network with UQ in a noisier environment. We first measure the activity detection accuracy and F-score of the $\alpha$-$\beta$ network, shown in Table \ref{tab:AccuracyFscore_inhouse}. \textcolor{black}{It can be seen that our method still outperforms the baseline (84\% vs. 80\%). Note that our model did drop accuracy compared to the average results from Table \ref{tab:AccuracyFscore} (84.3\% (std 0.015)), showing we indeed collected a noisier dataset which is more realistic. Secondly,} to present the model's ability to detect new contexts on such a dataset, we present the UQ results as well, shown in Figure  \ref{fig:inhouse_UQ}. It is clear that unknown contexts are more spread out, whereas known contexts have more concentrated predictive posterior probability distributions. We have further pushed the experimental setting harder to make it more realistic. That is, \textcolor{black}{in this dataset with three contexts} we removed 2 contexts as unknown contexts and just run the UQ solely on the seen contexts, and calculate the probability distribution to see if it can distinguish different contexts. As a result, the model can still find concentrated probability from known contexts but flat distribution for unknown ones, showing our superior performance on unknown context discovery.

\begin{table}[t!]
\caption{In house data, Accuracy and F score for the $\alpha$-$\beta$ network and baseline which is a $\beta$ network with the same number of parameters for each specific number of clusters. For $1$ cluster both the models are the same.}
\label{tab:AccuracyFscore_inhouse}
\vskip 0.15in
\begin{center}
\begin{normalsize}
\begin{tabular}{ccc}
\hline
Performance Measures    & Accuracy & F score\\
\hline
$\alpha$-$\beta$ Network & 0.84 & 0.86 \\
baseline & 0.80 & 0.80 \\
\hline
\end{tabular}
\end{normalsize}
\end{center}
\vskip -0.1in
\end{table}

\section{Conclusion}\label{conclusion}

In this paper, we developed a novel $\alpha$-$\beta$ network together with its UQ formulation in tackling a range of realistic situations where context information was unknown but critical for enhanced situation awareness and human activity recognition. Experiments on a real-world dataset showed that this combination of deep learning and uncertainty quantification led to superior performances by its efficacy and efficiency in extracting context information, recognizing unknown contexts, and assessing prediction's uncertainty. 

This work can be used as a foundation for a more comprehensive analysis of contextual discovery in a variety of modeling efforts in the future. Multiple levels of contextual information can be gathered and learned from the system, and understanding those in a truly unsupervised setting can enhance a number of recognition tasks and create a flexible and more realistic ontology for how to define context in human activity recognition tasks, rather than relying on high-level but general descriptions of context or too restrictive pre-defined contexts. In addition, other sources of uncertainty in prediction which result from data of new distributions or noisy data from old distributions should be considered as well.

\subsection*{Acknowledgements}
This project is in part supported by the Defense Advanced Research Projects Agency under grand FA8750-18-2-0027 and National Institutes of Health under grant 1R01EB028106-01 and 1R21EB028486-01.

\bibliography{bib}

\end{document}